\def\year{2022}\relax
\documentclass[letterpaper]{article} 
\usepackage{aaai22}  
\usepackage{times}  
\usepackage{helvet}  
\usepackage{courier}  
\usepackage[hyphens]{url}  
\usepackage{graphicx} 
\usepackage{natbib}  
\usepackage{caption} 
\DeclareCaptionStyle{ruled}{labelfont=normalfont,labelsep=colon,strut=off} 
\usepackage{algorithm}

\usepackage{newfloat}
\usepackage{listings}
\floatstyle{ruled}
\newfloat{listing}{tb}{lst}{}
\floatname{listing}{Listing}

\usepackage{xspace}
\makeatletter
\DeclareRobustCommand\onedot{\futurelet\@let@token\@onedot}
\def\@onedot{\ifx\@let@token.\else.\null\fi\xspace}
\def\eg{\emph{e.g}\onedot} 
\def\ie{\emph{i.e}\onedot} 
 
\def\etc{\emph{etc}\onedot}

\newcommand{\expectation}{\mathop{\mathbb{E}}}

\usepackage{amsmath}
\usepackage{amssymb}
\usepackage{color}

\usepackage{multirow}
\usepackage{booktabs}
\usepackage{makecell}
\usepackage[table,dvipsnames,svgnames,x11names]{xcolor}

\DeclareMathOperator*{\argmax}{arg\,max}

\definecolor{darkgreen}{RGB}{30,150,30}
\definecolor{darkblue}{RGB}{0,0,127}
\definecolor{darkyellow}{RGB}{171,133,0}
\definecolor{darkred}{RGB}{180,20,20}
\definecolor{darkmagenta}{RGB}{127,0,127}
\definecolor{darkcyan}{RGB}{0,127,127}

\definecolor{purple}{HTML}{9900ff}
\definecolor{darkpink}{HTML}{ff00ff}
\definecolor{maroon}{HTML}{980000}

\newcommand{\ak}[1]{\textcolor{red}{#1}} 

\definecolor{coolblack}{rgb}{0.0, 0.23, 0.64}
\newcommand{\jnkc}[1]{\textcolor{coolblack}{#1}} 

\usepackage{algpseudocode}

\algnewcommand\algorithmicforeach{\textbf{for each}}
\algdef{S}[FOR]{ForEach}[1]{\algorithmicforeach\ #1\ \algorithmicdo}

\usepackage{array}
\newcolumntype{P}[1]{>{\centering\arraybackslash}p{#1}}

\algnewcommand{\algorithmicgoto}{\textbf{go to}}%
\algnewcommand{\Goto}[1]{\algorithmicgoto~\ref{#1}}%

\usepackage{pict2e,picture,graphicx}
\makeatletter
\DeclareRobustCommand{\Arrow}[1][]{%
\check@mathfonts
\if\relax\detokenize{#1}\relax
\settowidth{\dimen@}{$\m@th\rightarrow$}%
\else
\setlength{\dimen@}{#1}%
\fi
\sbox\z@{\usefont{U}{lasy}{m}{n}\symbol{41}}%
\begin{picture}(\dimen@,\ht\z@)
\roundcap
\put(\dimexpr\dimen@-.7\wd\z@,0){\usebox\z@}
\put(0,\fontdimen22\textfont2){\line(1,0){\dimen@}}
\end{picture}%
}
\makeatother
\newcommand{\shortarrow}{\hspace{.2mm}\scalebox{.8}{\Arrow[.1cm]}\hspace{.2mm}}

\usepackage[switch]{lineno}

\usepackage[pagebackref=true,breaklinks=true,letterpaper=true,colorlinks,bookmarks=false]{hyperref}
\hypersetup{
     citecolor = coolblack,
}

\title{Amplitude Spectrum Transformation for \\ Open Compound Domain Adaptive Semantic Segmentation}
\author {
    Jogendra Nath Kundu\textsuperscript{\rm 1}\thanks{equal contribution}, Akshay Kulkarni\textsuperscript{\rm 1}\footnotemark[1], Suvaansh Bhambri\textsuperscript{\rm 1}\footnotemark[1], \\
    Varun Jampani\textsuperscript{\rm 2}, R. Venkatesh Babu\textsuperscript{\rm 1}
}
\affiliations {
    \textsuperscript{\rm 1}Indian Institute of Science, Bangalore \;\;\;
    \textsuperscript{\rm 2}Google Research
}

\begin{document}

\maketitle

\begin{abstract}

Open compound domain adaptation (OCDA) has emerged as a practical adaptation setting which considers a single labeled source domain against a compound of multi-modal unlabeled target data in order to generalize better on novel unseen domains. We hypothesize that an improved disentanglement of domain-related and task-related factors of dense intermediate layer features can greatly aid OCDA. Prior-arts attempt this indirectly by employing adversarial domain discriminators on the spatial CNN output. However, we find that latent features derived from the Fourier-based amplitude spectrum of deep CNN features hold a more tractable mapping with domain discrimination. Motivated by this, we propose a novel feature space Amplitude Spectrum Transformation (AST)\footnote{Project page: \url{https://sites.google.com/view/ast-ocdaseg}}. 
During adaptation, we employ the AST auto-encoder for two purposes. First, carefully mined source-target instance pairs undergo a simulation of cross-domain feature stylization (AST-Sim) at a particular layer by altering the AST-latent. Second, AST operating at a later layer is tasked to normalize (AST-Norm) the domain content by fixing its latent to a mean prototype. 
Our simplified adaptation technique is not only clustering-free but also free from complex adversarial alignment. We achieve leading performance against the prior arts on the OCDA scene segmentation benchmarks.


\end{abstract}

%
\vspace{-2mm}
\section{Introduction}
Deep learning has shown unprecedented success in the challenging semantic segmentation task (\jnkc{Long et al.} \citeyear{long2015fully}). In a fully supervised setting \cite{chen2018encoder,kundu2020vrt}, most approaches operate under the assumption that the training and testing data are drawn from the same input distribution. Though these approaches
work well on several benchmarks like 
Cityscapes \cite{cordts2016cityscapes}, their poor generalization to unseen datasets is often argued as the major shortcoming \cite{hoffman2016fcns}. Upon deployment in real-world settings, they fail to replicate the benchmark performance. This is attributed to the discrepancy in input distributions or domain-shift (\jnkc{Torralba et al.} \citeyear{torralba2011unbiased}). A naive solution would be to annotate the target domain samples. However, huge cost of annotation and variety of distribution shifts that could be encountered in future render this infeasible. Addressing this, unsupervised domain adaptation (DA) has emerged as a suitable problem setup, that aims to transfer the knowledge from a labeled source domain to an unlabeled target domain.


In recent years, several unsupervised DA techniques for semantic segmentation have emerged. Such as, techniques inspired from adversarial alignment \cite{tsai2018learning,kundu2019umadapt,kundu2018adadepth}, style transfer \cite{hoffman2018cycada}, pseudo-label self-training \cite{zou2019confidence}, \etc. However, these methods assume the target domain to be a single distribution. This assumption is difficult to satisfy in practice, as test images may come from mixed or continually changing or even unseen conditions. For example, data for self-driving applications may be collected from different weather conditions \cite{SDHV18} or different time of day (\jnkc{Sakaridis et al.} \citeyear{SDV20}) or different cities \cite{chen2017no}.
Towards a realistic DA setting, \citet{liu2020open} introduced Open Compound DA (OCDA) by incorporating mixed domains (\textit{compound}) in the target but without domain labels. Further, \textit{open} domains are available only for evaluation, representing unseen domains.

The general trend in OCDA (see Table \ref{tab:characteristics}) has been to break down the complex problem into multiple easier single-target DA problems and employ variants of existing unimodal DA approaches. To enable such a breakdown, \citet{chen2019blendingtarget} rely on unsupervised clustering to obtain sub-target clusters. Post clustering, both DHA \cite{park2020discover} and MOCDA \cite{gong2021cluster} embrace domain-specific learning. DHA employs separate discriminators and MOCDA uses separate batch-norm parameters (for each sub-target cluster), followed by complex adversarial alignment training for both. Though such domain-specific learning seems beneficial for compound domains, it hurts the generalization to open domains. To combat this generalization issue, MOCDA proposes online model update on encountering the open domains. Note that, such extra updates impedes the deployment-friendliness.
In this work, our prime objective is to devise a simple and effective OCDA technique. To this end, we aim to eliminate the requirement of \textbf{a)} adversarial alignment, \textbf{b)} sub-target clustering, \textbf{c)} incorporating domain-specific components, and \textbf{d)} online model update (refer the last row of Table \ref{tab:characteristics}).    



\newcolumntype{a}{>{\columncolor{gray!10}}c}
\begin{table}[t]
\centering
\caption{
    Characteristic comparison against prior OCDA works. 
    Note that \textit{cluster} refers to the {sub-target} clusters, \textit{I2I n/w} indicates image-to-image translation network and \textit{Disc.} indicates use of adversarial discriminator.
}
\vspace{-3mm}
\label{tab:characteristics}
\setlength{\tabcolsep}{0.01pt}
\resizebox{1\columnwidth}{!}{
\begin{tabular}{lacacac}
    \toprule
        & \begin{tabular}{c} Complex \\ adv. training \\ (via Disc.) \end{tabular}  & \begin{tabular}{c} Model update \\  for open \\ domains \end{tabular}  & \begin{tabular}{c} Explicit \\domain-specific\\ learning \end{tabular} & Clustering & \begin{tabular}{c}Extra \\networks\end{tabular}   \\
        \midrule
        \begin{tabular}{l}OCDA \\{\small\cite{liu2020open}}\end{tabular} & $\checkmark$ & $\times$ & $\times$ & $\times$ & Disc. \\
        \hline
        \begin{tabular}{l}DHA\\ {\small\cite{park2020discover}}\end{tabular} & $\checkmark$ & $\times$ & \begin{tabular}{c}$\checkmark$ (Separate Disc. \\ for each cluster)\end{tabular} & $\checkmark$ & \begin{tabular}{c}I2I n/w, \\Disc.\end{tabular} \\
        \hline
        \begin{tabular}{l}MOCDA\\ {\small\cite{gong2021cluster}}\end{tabular} & $\checkmark$ & \begin{tabular}{c}$\checkmark$ \\ (online update)\end{tabular} & \begin{tabular}{c}$\checkmark$ (Separate BN \\ for each cluster)\end{tabular} & $\checkmark$ & \begin{tabular}{c} Hyper n/w, \\Disc.\end{tabular} \\
        \hline
        Ours & $\times$ & $\times$ & $\times$ & $\times$ & AST \\
    \bottomrule
\end{tabular} \vspace{-2mm}
} 
\end{table}

To this end, we uncover key insights while exploring along the lines of disentangling domain-related and task-related cues at different layers of the segmentation architecture. We perform control experiments to quantify the unwanted correlation of the deep features with the unexposed domain labels by introducing a domain-discriminability metric (DDM). DDM accuracy indicates the ease of classifying the domain label of deep features at different layers. We observe that deeper layers hold more domain-specificity and identify this as a major contributing factor to poor generalization. To alleviate this, prior-arts \cite{park2020discover} employ domain discriminators on the spatial deep features. We ask, can we get hold of a representation space that favors domain discriminability better than raw spatial deep features? Being able to do so would provide us better control to manipulate the representation space for effective adaptation.

To this end, we draw motivation from the recent surge in the use of frequency spectrum analysis to aid domain adaptation \cite{yang2020fda, yang2020phase, huang2021fsdr}. These approaches employ different forms of Fourier transform (FFT) to separately process the phase and amplitude components in order to carry out content-preserving image augmentations. Motivated by this, we propose to use a latent neural network mapping derived from the amplitude spectrum of the raw deep features as the desired representation space. Towards this, we develop a novel feature-space Amplitude Spectrum Transformation (AST) to auto-encode the amplitude while preserving the phase. We empirically confirm that the DDM of AST-latent is consistently higher than the same for the corresponding raw deep features. To the best of our knowledge, we are the first to use frequency spectrum analysis on spatial deep features. 

The AST-latent lays a suitable ground to better discern the domain-related factors which in turn allows us to manipulate the same to aid OCDA. We propose to manipulate AST-latent in two ways. First, carefully mined source-target instance pairs undergo a simulation of cross-domain feature stylization (\textit{AST-Sim}) at a particular layer by altering the AST-latent. Second, AST operating at a later layer is tasked to normalize (\textit{AST-Norm}) the domain content by fixing its latent to a mean prototype. As the phase is preserved in AST, both the simulated and domain normalized features retain the task-related content while altering the domain-related style. However, the effectiveness of this intervention is proportional to the quality of disentanglement. Considering DDM as a proxy for disentanglement quality, we propose to apply \textit{AST-Sim} at a deeper layer with high DDM followed by applying AST-Norm at a later layer close to the final task output. A post-adaptation DDM analysis confirms that the proposed \textit{Simulate-then-Normalize} strategy is effective enough to suppress the domain-discriminability of the deeper features, thus attaining improved domain generalization compared to typical adversarial alignment. We summarize our contributions below:

\begin{itemize}
\item We propose a novel feature-space Amplitude Spectrum Transformation (AST), based on a thorough analysis of domain discriminability, for improved disentanglement and manipulability of domain characteristics.

\item We provide insights into the usage of AST in two ways - \textit{AST-Sim} and \textit{AST-Norm}, and propose a novel \textit{Simulate-then-Normalize} strategy for effective OCDA.

\item Our proposed approach achieves \textit{state-of-the-art} semantic segmentation performance on GTA5$\to$C-Driving and SYNTHIA$\to$C-Driving benchmarks for both compound and open domains, as well as on generalization to extended open domains Cityscapes, KITTI and WildDash.

\end{itemize}


\vspace{-2mm}
\section{Related Works}
\label{sec:related-works}

\noindent
\textbf{Open Compound Domain Adaptation.} 
\citet{liu2020open} proposes to improve the generalization ability of the model on compound and open domains by using a memory based curriculum learning approach. 
DHA \cite{park2020discover} and MOCDA \cite{gong2021cluster} cluster the target using $k$-means clustering on convolutional feature statistics and encoder features of image-to-image translation network respectively. DHA performs adversarial alignment between the source and each sub-target cluster while MOCDA uses separate batch norm parameters for each sub-target cluster with a single adversarial discriminator.
In contrast, we simulate the target style by manipulating the amplitude spectrum of our source features in the latent space.

\vspace{1mm}
\noindent \textbf{Stylization for DA.}
Several recent works that use stylization for DA can be broadly divided into - feature-statistics-based and FFT-based.
First, the properties of Fourier transform (FFT), \ie disentangling the input into phase spectrum (representing content) and amplitude spectrum (representing style), made it a natural choice for recent DA works \cite{yang2020fda}. Several prior arts \cite{yang2020label, kundu2021generalize} employ the FFT directly on the input RGB image in order to simulate some form of image-to-image domain translation to aid the adaptation. 
In contrast to these works, we utilize the FFT on the CNN feature space. 
Second, the feature-statistics-based methods \cite{kim2020learning} use the first or second-order convolutional feature statistics to perform feature-space stylization \cite{huang2017adain} followed by reconstructing the stylized image via a decoder. The stylized images are used with other adaptation techniques like adversarial alignment.
Contrary to this, we perform content-preserved feature-space stylization using the Amplitude Spectrum Transformation (AST).
\vspace{-3mm}
\section{Approach}

In this section, we thoroughly analyze domain discriminability (Sec \ref{subsec:disentangling}). Based on our observations, we propose the Amplitude Spectrum Transformation (Sec \ref{subsec:ast}) and provide insights for effectively tackling OCDA (Sec \ref{subsec:ocda_training}).


\begin{figure*}[t]
\centering
\vspace{-4mm}
\includegraphics[width=\textwidth]{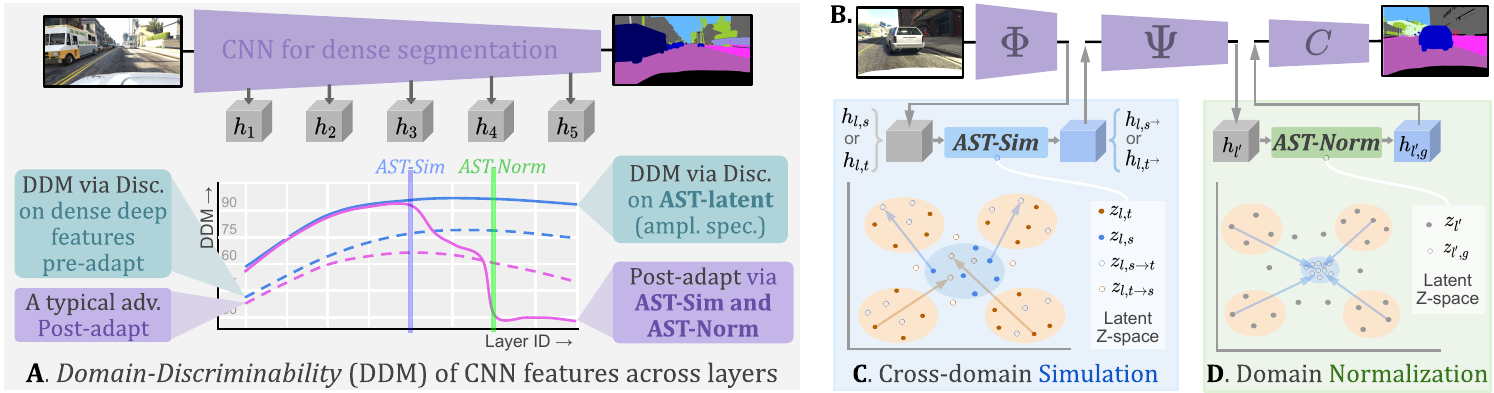} 
\caption{\textbf{A.} 
DDM computed on the AST-latent is consistently higher than the same on raw deep features (Sec \ref{subsec:disentangling}). The post-adaptation DDM confirms the effectiveness of the proposed \textit{Simulate-then-Normalize} strategy (Sec \ref{subsec:simthennorm}). \textbf{B.} Integration of AST-Sim and AST-Norm into the CNN segmentor (Sec \ref{subsec:simthennorm}\ak{.1}). \textbf{C.} \& \textbf{D.} Internal working of AST-Sim and AST-Norm (Sec \ref{subsec:ocda_training}).
}\vspace{-1mm}
\label{fig:observation_approach}
\end{figure*}

\subsection{Disentangling domain characteristics}
\label{subsec:disentangling}



In the literature of causal analysis \cite{achille2018emergence}, it has been shown that generally trained networks ({ERM-networks}) are prone to learn domain-specific spurious correlations. Here, {ERM-network} refers to a model trained via empirical risk minimization (ERM) on multi-domain mixture data (without the domain label). In order to test the above proposition, we introduce a metric that would capture the unwanted correlation of layer-wise deep features with the unexposed domain labels. 
A higher correlation indicates that the model inherently distinguishes among samples from different domains and extracts domain-specific features followed by domain-specific hypothesis (mapping from input to output space) learning. To this end, we introduce the following metric as a proportional measure of this correlation.

\vspace{1mm}
\noindent
\textbf{3.1.1 Quantifying domain discriminability (DDM).} 
Domain-Discriminability Metric (DDM) 
is an accuracy metric that measures the discriminativeness of the features for domain classification. Given a CNN segmentor (see Fig.~\ref{fig:observation_approach}\ak{A}), $\{h_k\}_{k=1}^K$ denote the 3D tensor of neural activities at the output of convolutional layers with $k$ as the layer depth. Here, $h_k\!\in\!\mathcal{H}_k$ with $\mathcal{H}_k$ denoting the raw spatial CNN feature space. Following this, DDM is denoted as $\lambda_k$.

\vspace{0.5mm}
\noindent
\textbf{Empirical analysis.} We obtain a multi-domain source variant with 4 sub-domains by performing specific domain-varying image augmentations such as weather augmentation, {cartoonization}, \etc (refer Suppl.). Following this, a straightforward procedure to compute DDM for each layer $k$ would be to report the accuracy of fully-trained domain discriminators operating on the intermediate deep features of the frozen ERM-network. 
In Fig. \ref{fig:observation_approach}\ak{A}, the dashed blue curve shows the same in a plot. A peculiar observation is that DDM increases while traversing from input towards output.

\vspace{1mm}
\noindent \textbf{Observation 1.} \textit{An ERM-network trained on multi-domain data for dense semantic segmentation tends to learn increasingly more domain-specific features, in the deeper layers.}

\vspace{1mm}
\noindent \textbf{Remarks.}
This is in line with the observation of increasing memorization in the deeper layers against the early layers as studied by \citet{stephenson2021on}. This is because the increase in feature dimensions for deeper layers allows more room to learn unregularized domain-specific hypotheses. This effect is predominant for dense prediction tasks \cite{kundu2020kinematicstructurepreserved}, as domain-related information is also retained as higher-order clues along the spatial dimensions, implying more room to accommodate domain-specificity. A pool of OCDA or UDA methods \cite{park2020discover, tsai2018learning}, that employ adversarial domain alignment, aim to minimize the DDM of deeper layer features as a major part of the adaptation process (dashed magenta curve in Fig. \ref{fig:observation_approach}\ak{A}).


\vspace{1mm}
\noindent
\textbf{3.1.2 \hspace{1mm}Seeking latent representation for improved DDM.}

\noindent
We ask ourselves, can we get hold of a representation space that favors domain discriminability better than the raw spatial deep features? Let, $\mathcal{Z}_k$ be a latent representation space where the multi-domain samples are easily separable based on their domain label. We expect the DDM value to increase if the corresponding discriminator operates on this latent representation instead of the raw spatial deep features. Essentially, we seek to learn a mapping $\mathcal{H}_k\!\to\!\mathcal{Z}_k$ which would encourage a better disentanglement of domain-related factors from the spatial deep features which are known to hold entangled domain-related and task-related information. 

To this end, we draw motivation from the recent surge in the use of frequency spectrum analysis to aid domain adaptation \cite{yang2020fda, yang2020label, huang2021fsdr}. These approaches employ different forms of Fourier transform (FFT) to separately process the phase and amplitude components 
in order to carry out content-preserving image augmentations. Though both amplitude and phase are necessary to reconstruct the spatial map, it is known that magnitude-only reconstructions corrupt the image making it unrecognizable while that is not the case for phase-only reconstructions. In other words, changes in the amplitude spectrum alter the global style or appearance, i.e., the domain-related factors while the phase holds onto the crucial task-related information. Based on these observations, we hypothesize that the domain-related latent, $\mathcal{Z}_k$ can be derived from the amplitude spectrum.

\vspace{0.5mm}
\noindent
\textbf{Empirical analysis.} Based on the aforementioned discussion, we develop a novel Amplitude Spectrum Transformation (AST). Similar to an auto-encoding setup, AST involves the encoding ($\mathcal{H}\!\to\!\mathcal{Z}$) and decoding ($\mathcal{Z}\!\to\!\mathcal{H}$) transformations operating on the amplitude spectrum of the feature maps. Refer Sec \ref{subsec:ast} for further details. Following this, DDM is obtained as the accuracy of a domain discriminator network operating on latent $\mathcal{Z}$ space (repeated for each layer $k$). In Fig. \ref{fig:observation_approach}\ak{A}, the solid blue curve shows the same in a plot.


\vspace{1mm}
\noindent
\textbf{Observation 2.} \textit{Domain discriminability (and thus DDM) is easily identifiable and manipulatable in the latent} $\mathcal{Z}_k$ \textit{space,} \ie\textit{a latent neural network mapping derived from the amplitude spectrum of the raw spatial deep features (}$\mathcal{H}_k$ \textit{space).}

\vspace{1mm}
\noindent
\textbf{Remarks.} 
Several prior works (\jnkc{Gatys et al.} \citeyear{gatys2016image}; \jnkc{Huang et al.} \citeyear{huang2017adain}) advocate that spatial feature statistics capture the style or domain-related factors. Following this, style transfer networks propose to normalize the channel-wise first or second-order statistics, \eg, instance normalization. One can relate the latent AST representation as a similar measure to represent complex domain discriminating clues that are difficult to extract via multi-layer convolutional discriminators.

\subsection{Amplitude Spectrum Transformation (AST)}
\label{subsec:ast}

Several prior arts (\jnkc{Yang et al.} \citeyear{yang2020fda}; \jnkc{Huang et al.} \citeyear{huang2021fsdr}) employ frequency spectrum analysis directly on the input RGB image to simulate some form of image-to-image domain translation. To the best of our knowledge, we are the first to use frequency spectrum analysis on spatial deep features.

\subsubsection{3.2.1 AST as an auto-encoding setup.}
\begin{figure}[t]
\centering
\vspace{-1mm}
\includegraphics[width=\columnwidth]{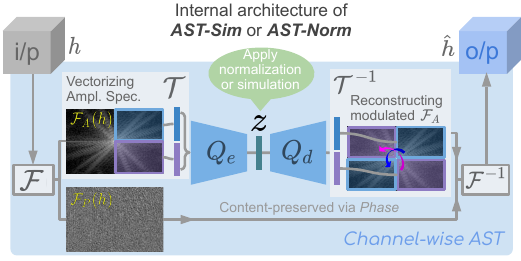} 
\vspace{-6mm}
\caption{Internal architecture of AST auto-encoder.}
\label{fig:AST_internal_arch}
\end{figure}

Broadly, AST involves an encoder mapping and a decoder mapping. As shown in Fig. \ref{fig:AST_internal_arch}, the encoder mapping involves a Fourier transform $\mathcal{F}$ of the input feature $h$ to obtain the amplitude spectrum $\mathcal{F}_A(h)$ and the phase spectrum $\mathcal{F}_P(h)$. The amplitude spectrum is passed through a transformation $\mathcal{T}$ and a fully-connected encoder network $Q_e$ to obtain the AST-latent $z = Q_e \circ \mathcal{T} \circ \mathcal{F}_A(h)$. Consequently, the decoder mapping involves the inverse transformations to obtain $\hat{h}=\mathcal{F}^{-1}(\mathcal{T}^{-1} \circ Q_d(z), \mathcal{F}_P(h))$. It is important to note that the encoding side phase spectrum is directly forwarded to the decoding side to be recombined with the reconstructed amplitude spectrum via inverse Fourier transform $\mathcal{F}^{-1}$.  

Here, the {vectorization ($\mathcal{T}$) and inverse-vectorization ($\mathcal{T}^{-1}$) of the amplitude spectrum} must adhere to its symmetric about origin (mid-point) property. To this end, $\mathcal{T}$ only vectorizes the first and fourth quadrants and $\mathcal{T}^{-1}$ reconstructs the full spectrum from the reshaped first and fourth quadrants by mirroring about the origin (see Fig. \ref{fig:AST_internal_arch}).



\textbf{Training.} Let $\theta_Q^{(k)}=\{\theta_{Q_e}, \theta_{Q_d}\}$ denote the trainable parameters of AST at the $k^\textit{th}$ layer. Note that, AST operates independently for each channel of $h_k$ with the same parameters. Effectively, $z_k$ is obtained as a concatenation of $z$'s from each channel of $h_k$. After obtaining the frozen \textit{ERM-network}, AST, at a given layer, is trained to auto-encode the amplitude spectrum using the following objective,
\begin{equation}
    \min_{\theta_Q} \mathcal{L}_\texttt{AST}(h_k, \hat{h}_k) \; \text{where} \; \mathcal{L}_\texttt{AST}(h_k, \hat{h}_k) = \Vert h_k - \hat{h}_k \Vert_2^2
    \label{eqn:AST_ae_training}
\end{equation}

For simplicity, we denote the overall AST as a function \texttt{AST} and simplify the reconstruction as $\hat{h}_k = \texttt{AST}(h_k, z_k)$. 
Note that we omit the $z_k$ term in the auto-encoding phase \ie $\hat{h}_k = \texttt{AST}^{(\textit{ae})}(h_k)$ as $z_k$ is passed through unmodified.

\subsection{Usage of AST for OCDA}
\label{subsec:ocda_training}

AST lays a suitable ground to better identify  domain-related factors (high DDM) allowing us to manipulate the same.

\vspace{0.5mm}
\noindent
\textbf{Insight 1.} \textit{Altering the latent $z$ by the same obtained from a different domain or instance (}\ie \textit{$\tilde{z}$) while transforming $\tilde{h}\!=\!\texttt{AST}(h, \tilde{z})$ is expected to imitate feature space cross-domain translation ($h$-to-$\tilde{h}$) as the crucial task-related information remains preserved via the unaltered phase spectrum. }




\subsubsection{3.3.1 AST for cross-domain simulation (\textit{AST-Sim}).}
\label{subsubsec:crossmodalsim}
One of the major components in prior OCDA approaches is to discover sub-target clusters as it is used either for domain-translation of source samples \cite{park2020discover} or to realize the domain-specific batch-normalization \cite{gong2021cluster}. In order to obtain reliable sub-target clusters, both works rely on 
unsupervised clustering of the compound target data. In contrast, we aspire to avoid such discovery and do away with introducing an additional hyperparameter (no. of clusters) in order to realize a clustering-free OCDA algorithm. 

\textbf{Cross-domain pairing.} 
How to select reliable source-target pairs for cross-domain simulation? In the absence of sub-target clusters, a naive approach would be to form random pairs of source and target instances. Aspiring to formalize a better source-target pairing strategy, we propose to pair instance with the most distinct domain style. This is motivated from the hard negative mining as used in deep metric learning approaches \cite{suh2019stochastic}. 
%
%
%
Consider a source dataset with instances $\{s_i\}_{i=1}^{N_s}$ and a target dataset with instances $\{t_j\}_{j=1}^{N_t}$. For each target instance $t_j$, we mine a source instance $s_i$ such that these are maximally separated in the AST-latent space of the $l^\text{th}$ CNN layer. Formally, we obtain a set of cross-domain pairs as,

\vspace{-2mm}
\begin{equation}
    \mathcal{U} = \{ (s_i, t_j): \forall t_j, i = \arg\max_{i'} \zeta(z_{l, s_{i'}}, z_{l, t_j}) \}
\end{equation}

\noindent
where $z_{l, s_i}$ and $z_{l, t_j}$ are the $l^\text{th}$ layer AST-latent for the $s_i$ and $t_j$ instances respectively. Here, $\zeta$ is a L2 distance function. 

\textbf{Cross-domain simulation.} Utilizing Insight \ak{1}, we simulate the style of $t_j$ in the source $s_i$ by replacing its AST-latent \ie, 
%
    $\hat{h}_{l, s_{i\shortarrow j}} = \texttt{AST}(h_{l, s_i}, z_{l, t_j})$.
%
Similarly, $\hat{h}_{l, t_{j \shortarrow i}}$ represents the source stylized target feature. 
We illustrate the cross-domain simulation in Fig. \ref{fig:observation_approach}\ak{C}.
For notation simplification, $\hat{h}_{l, s_{i \shortarrow j}}$ and $\hat{h}_{l, s_{i \shortarrow j}}$ are denoted as $h_{l, s^{\shortarrow}}$ and $h_{l, t^{\shortarrow}}$ respectively. Note that, $h_{l, s^{\shortarrow}}$ and $h_{l, t^{\shortarrow}}$ hold the same task-related content as $h_{l, s}$ and $h_{l, t}$ respectively. Thus, both original and stylized features are expected to have the same segmentation label.


\vspace{0.5mm}
\noindent
\textbf{3.3.2 AST for domain normalization (\textit{AST-Norm}).}
\label{subsubsec:ast_dn}
We draw motivation from batch normalization (BN) (\jnkc{Ioffe et al.} \citeyear{ioffe2015batch}), which normalizes the features with first and second order statistics during forward pass. Along the same lines, we aim to normalize the AST-latent of intermediate deep features by altering it with a fixed mean prototype. 
%
%
Let $\mathcal{V}$ be the set of $l^{'\text{th}}$-layer deep features obtained from all the four variants, \ie, $h_{l'\!,s}, h_{l'\!,t}, h_{l'\!,s^{\shortarrow}}, h_{l'\!,t^{\shortarrow}}$. 
%
%
We compute the fixed domain prototype $z_{l'\!,g}$ as  follows, 
%
%
\begin{equation}
    z_{l'\!,g}=\expectation_{h \in \mathcal{V}} \; [Q_e \circ \mathcal{T} \circ \mathcal{F}_A(h)]
    \label{eqn:fixeddomainprototype}
\end{equation}

Following Insight \ak{1}, the domain normalization is performed as
$h_{l'\!, s_g} = \texttt{AST}(h_{l'\!, s}, z_{l'\!,g})$.
We illustrate the domain normalization in Fig. \ref{fig:observation_approach}\ak{D}.
The normalized versions of $h_{l'\!,t}, h_{l'\!,s^{\shortarrow}}, h_{l'\!,t^{\shortarrow}}$ are $h_{l'\!,t_g}, h_{l'\!,s^{\shortarrow}_g}, h_{l'\!,t^{\shortarrow}_g}$ respectively. 



\subsection{Simulate-then-Normalize for OCDA}
\label{subsec:simthennorm}
The prime objective in OCDA is to adapt the task-knowledge from a labeled source domain $(x_s, y_s) \in \mathcal{D}_s$ to an unlabeled compound target, $x_t \in \mathcal{D}_t$ towards better generalization to unseen open domains. 
%
%
%
%
%
The advantages of AST (\ie identifying and manipulating domain-related factors) well cater to the specific challenges encountered in OCDA against employing it for single-source single-target settings.

\vspace{0.5mm}
\noindent
\textbf{Empirical analysis.} After obtaining the frozen ERM-network and the frozen layer-wise AST modules, we employ \textit{AST-Sim} (Fig. \ref{fig:observation_approach}\ak{C}) and \textit{AST-Norm} (Fig. \ref{fig:observation_approach}\ak{D}) to analyze its behavior towards the task performance. We observe that performing \textit{AST-Sim} or \textit{AST-Norm} at the earlier layer with low DDM hurts the task performance. We infer that a low DDM value indicates inferior disentanglement of domain-related cues at the corresponding AST-latent. In other words, the task-related cues tend to leak through the amplitude spectrum thereby distorting task-related content.

\vspace{1mm}
\noindent
\textbf{Insight 2.} \textit{Applying \textit{AST-Sim} at a layer with a high DDM value followed by applying \textit{AST-Norm} at a later layer close to the final task output helps to effectively leverage the advantages of AST modules for the challenging OCDA setting.}

\vspace{1mm}
\noindent
\textbf{Remarks.} From the perspective of causal analysis literature, the use of \textit{AST-Sim} can be thought of as an intervention (\jnkc{He et al.} \citeyear{he2008active}) that implicitly encourages the network to focus on the uninterrupted task-specific cues. However, the effectiveness of this intervention is proportional to the quality of disentanglement. Here, DDM can be taken as a proxy to measure the degree of disentanglement quality. Thus, we propose to perform \textit{AST-Sim} at a layer $l$, with high DDM, as indicated in Fig. \ref{fig:observation_approach}\ak{B}. However, we realize that the uninterrupted phase pathway still leaks domain-related information (relatively high DDM 
for $k>l$) hindering domain generalization. To circumvent this, we propose to perform \textit{AST-Norm} at a later layer as a means to further fade away the domain-related information. A post-adaptation DDM analysis (solid magenta curve in Fig. \ref{fig:observation_approach}\ak{A}) confirms that the proposed \textit{Simulate-then-Normalize} strategy is effective enough to suppress the domain-discriminability of the later deep features, thus attaining improved domain generalization.

\vspace{0.5mm}
\noindent
\textbf{Architecture.} As shown in Fig. \ref{fig:observation_approach}\ak{B}, we divide the CNN segmentor into three parts; \ie $\Phi, \Psi$, and $C$. Following this, the \textit{AST-Sim} (denoted as $\texttt{AST}_\textit{cs}$) and \textit{AST-Norm} (denoted as $\texttt{AST}_\textit{dn}$) are inserted after $\Phi$ and $\Psi$ respectively (see Fig. \ref{fig:detailed_flow}\ak{B}, \ak{D}). Next, we discuss the adaptation training.


\begin{algorithm}[!t]

\caption{Pseudo-code for the proposed approach}
\label{algo:overall}
\begin{algorithmic}[1]
\State \textbf{Input:} source data $(x_s, y_\textit{gt})\!\in\!\mathcal{D}_s$, target data $x_t \!\in\! \mathcal{D}_t$. Initialize $\theta$ (\ie parameters of $\Phi, \Psi$, and $C$) from standard source training. Initialize $\theta_\textit{cs}$ and $\theta_\textit{dn}$ (\ie parameters of $\texttt{ASM}_\textit{cs}$ and $\texttt{ASM}_\textit{dn}$) from auto-encoder training discussed in Sec \ref{subsec:ast}\ak{.1}. 

\State \underline{{\textit{Pre-adaptation}:}}
Integrate ASTs into the CNN segmentor and finetune $\theta$, $\theta_\textit{cs}$, and $\theta_\textit{dn}$ following Eq. \ref{eqn:sup_src_seg} and \ref{eqn:sup_ae}.


\vspace{1mm}

\Statex \underline{{\textit{Adaptation via Simulate-then-Normalize}}} \vspace{1mm}
\State Freeze $\theta_\textit{cs}$, and $\theta_\textit{dn}$ from the pre-adaptation training. Recompute fixed domain prototype $z_{l'\!, g}$ using Eq. \ref{eqn:fixeddomainprototype} and pseudo-labels $y_\textit{pgt}$ using Eq. \ref{eqn:pseudolabelextraction}, after every epoch. 

\For{$iter < MaxIter$}:
    \State $x_s, y_\textit{gt}, x_t, y_\textit{pgt} \leftarrow$ batch sampled from $\mathcal{D}_s, \mathcal{D}_t$
    \State Obtain predictions following Fig. \ref{fig:detailed_flow}
    \State \textbf{update} $\theta$ by optimizing Eq. \ref{eqn:srcsideadapt} and \ref{eqn:tgtsideadapt}
\EndFor
\end{algorithmic}
\end{algorithm}

\subsubsection{3.4.1 Pre-adaptation training.} In this stage, we aim to prepare (or initialize) the network components prior to commencing the proposed \textit{Simulate-then-Normalize} procedure (\ie adaptation). We start from the pre-trained standard source model and the pre-trained auto-encoder based AST-networks at layer $l$ and $l'$, \ie $\texttt{AST}_\textit{cs}^\textit{(ae)}$, and $\texttt{AST}_\textit{dn}^\textit{(ae)}$. Here, the superscript $\textit{(ae)}$ indicates that these networks are not yet manipulated as \textit{AST-Sim}, or \textit{AST-Norm} (\ie, without any manipulation of the AST-latent). 

Following this, the pre-adaptation finetuning involves two pathways. First, $x_s$ is forwarded to obtain 
$y_s = C\circ \texttt{AST}_\textit{dn}^\textit{(ae)}\circ\Psi\circ\texttt{AST}_\textit{cs}^\textit{(ae)}\circ \Phi(x_s)$. 
The following source-side supervised cross-entropy objective is formed as;
\begin{equation}
    \min_{\theta}\mathcal{L}_s\;\;\text{where}\;\; \mathcal{L}_s\!=\!L_\textit{CE}(y_s, y_\textit{gt})
    \label{eqn:sup_src_seg}
\end{equation}

Here, $\theta$ denotes the collection of parameters of $\Phi$, $\Psi$, and $C$. Note that, after integrating ASTs into the CNN segmentor, the content-related gradients flow unobstructed through the frozen AST networks while finetuning $\Phi$ and $\Psi$.



In the second pathway, both $x_s$ and $x_t$ are forwarded to obtain input-output pairs to finetune $\theta_\textit{cs}$, and $\theta_\textit{dn}$, \ie
\vspace{-2mm}
\begin{equation}
\begin{split}
    \min_{\theta_\textit{cs}}\mathcal{L}_\texttt{AST}(h_{l, s},\hat{h}_{l, s}) + \mathcal{L}_\texttt{AST}(h_{l, t},\hat{h}_{l, t})\\ 
    \min_{\theta_\textit{dn}}\mathcal{L}_\texttt{AST}(h_{l'\!,s},\hat{h}_{l'\!,s}) + \mathcal{L}_\texttt{AST}(h_{l'\!,t},\hat{h}_{l'\!,t})
    \end{split}
    \label{eqn:sup_ae}
\end{equation}

Here, $(h_{l, s},\hat{h}_{l, s})$ and $(h_{l, t},\hat{h}_{l, t})$ are the input-output pairs corresponding to $x_s$ and $x_t$ to update $\texttt{AST}_\textit{cs}^\textit{(ae)}$. Similarly, $(h_{l'\!,s},\hat{h}_{l'\!,s})$ and $(h_{l'\!,t},\hat{h}_{l'\!,t})$ are the input-output pairs corresponding to $x_s$ and $x_t$ to update $\texttt{AST}_\textit{dn}^\textit{(ae)}$.

\subsubsection{3.4.2 Adaptation via \textit{Simulate-then-Normalize}.}
The finetuned networks from the pre-adaptation stage are now subjected to adaptation via the simulate-then-normalize procedure. We manipulate the AST-latent thus denoting $\texttt{AST}_\textit{cs}^\textit{(ae)}$, and $\texttt{AST}_\textit{dn}^\textit{(ae)}$ as $\texttt{AST}_\textit{cs}$, and $\texttt{AST}_\textit{dn}$ respectively. Note that, latent manipulation is independent of the network weights, $\theta_\textit{cs}$, and $\theta_\textit{dn}$. Thus, we freeze $\theta_\textit{cs}$ and $\theta_\textit{dn}$ from the pre-adaptation training to preserve their auto-encoding behaviour, and only update $\theta$ during the adaptation stage.

\begin{figure}[t]
\centering
\vspace{-2mm}
\includegraphics[width=\columnwidth]{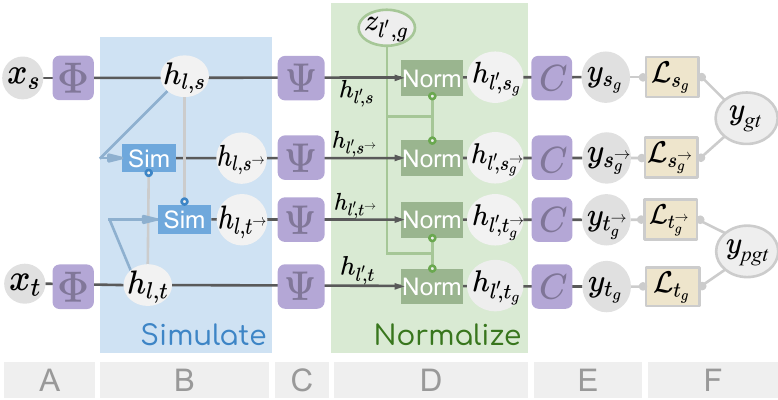} 
\vspace{-5mm}
\caption{Data flow during adaptation training (Sec \ref{subsec:simthennorm}\ak{.2}).}
\label{fig:detailed_flow}
\vspace{-4mm}
\end{figure}

The adaptation training involves four data-flow pathways as shown in Fig \ref{fig:detailed_flow}. The \textit{Simulate} step at layer-$l$ outputs the following features; $(h_{l, s}, h_{l, s^{\shortarrow}}, h_{l, t^{\shortarrow}}, h_{l, t})$. Moving forward, the \textit{Normalize} step at layer-$l'$ outputs the following: $(h_{l'\!, s_g}, h_{l'\!, s_g^{\shortarrow}}, h_{l'\!, t_g^{\shortarrow}}, h_{l'\!, t_g})$. Finally, we obtain the following four predictions; $(y_{s_g}, y_{s_g^{\shortarrow}}, y_{t_g^{\shortarrow}}, y_{t_g})$.



As shown in Fig. \ref{fig:detailed_flow}\ak{F}, we apply supervised losses against ground-truth (GT) $y_\textit{gt}$ for the source-side predictions \ie

\vspace{-4mm}
\begin{equation}
    \min_{\theta}(\mathcal{L}_{s_g}\texttt{=}L_\textit{CE}(y_{s_g}, y_\textit{gt}))\,\,\min_{\theta}(\mathcal{L}_{s_g^{\shortarrow}}\texttt{=}L_\textit{CE}(y_{s_g^{\shortarrow}}, y_\textit{gt}))
    \label{eqn:srcsideadapt}
\end{equation}

Following the self-training based adaptation works \cite{zou2019confidence}, we propose to use pseudo-labels (pseudo-GT) $y_\textit{pgt}$ for the target-side predictions \ie

\vspace{-4mm}
\begin{equation}
    \min_{\theta}(\mathcal{L}_{t_g}\texttt{=}L_\textit{CE}(y_{t_g}, y_\textit{pgt}))\,\,\min_{\theta}(\mathcal{L}_{t_g^{\shortarrow}}\texttt{=}L_\textit{CE}(y_{t_g^{\shortarrow}}, y_\textit{pgt}))
    \label{eqn:tgtsideadapt}
\end{equation}



\textbf{Pseudo-label extraction.} 
In order to obtain reliable pseudo-labels $y_\textit{pgt}$, we prune the target predictions by performing a consistency check between $y_{t_g}$ and $y_{t_g^{\shortarrow}}$. 
Thus, $y_\textit{pgt} = \argmax_c y_{t_g}$ is obtained only for reliable pixels \ie for pixels $x_t$ which satisfy the following criteria,
\begin{equation}
\mathcal{D}^\textit{(pgt)}_t \!=\! \{x_t\!:\! x_t\!\in\!\mathcal{D}_t \text{ and } \argmax_c y_{t_g} \!=\! \argmax_c y_{t_g^{\shortarrow}}\}
\label{eqn:pseudolabelextraction}
\end{equation}

\noindent
Note that, only the target pixels in $\mathcal{D}^\textit{(pgt)}_t$ contribute in Eq.~\ref{eqn:tgtsideadapt}. 
The overall training procedure is summarized in Algo. \ref{algo:overall}.




\begin{table}[t]
\centering
\vspace{-2mm}
\caption{
    Comparison on GTA5$\to$C-Driving benchmark. $\dagger$ indicates 150k training iterations, otherwise 5k iterations.
}
\vspace{-2mm}
\label{tab:gta5tocdriving}
\setlength{\tabcolsep}{1pt}
 \resizebox{1\columnwidth}{!}{
\begin{tabular}{lcccccc}
    \toprule
         \multirow{2}{*}{Method} & \multicolumn{3}{c}{Compound (C)} & \multicolumn{1}{c}{Open (O)} & \multicolumn{2}{c}{Average} \\
        \cmidrule(l{4pt}r{4pt}){2-4} \cmidrule(l{4pt}r{4pt}){5-5} \cmidrule(l{4pt}r{4pt}){6-7}
         & Rainy & Snowy & Cloudy & Overcast & C & C+O \\
        \midrule
        Source only & 16.2 & 18.0 & 20.9 & 21.2 & 18.9 & 19.1 \\
        ASN \cite{tsai2018learning} & 20.2 & 21.2 & 23.8 & 25.1 & 22.1 & 22.5 \\
        CBST \cite{zou2018unsupervised} & 21.3 & 20.6 & 23.9 & 24.7 & 22.2 & 22.6 \\
        IBN-Net \cite{pan2018IBN} & 20.6 & 21.9 & 26.1 & 25.5 & 22.8 & 23.5 \\
        PyCDA \cite{lian2019constructing} & 21.7 & 22.3 & 25.9 & 25.4 & 23.3 & 23.8 \\
        OCDA \cite{liu2020open} & 22.0 & 22.9 & 27.0 & 27.9 & 24.5 & 25.0 \\
        DHA \cite{park2020discover} & 27.0 & 26.3 & 30.7 & 32.8 & 28.5 & 29.2 \\
        \rowcolor{gray!10} \textit{Ours (AST-OCDA)} & 28.2 & 27.8 & 31.6 & 34.0 & \textbf{29.2} & \textbf{30.4} \\
        \midrule
        Source only$\dagger$ & 23.3 & 24.0 & 28.2 & 30.2 & 25.7 & 26.4 \\
        ASN \cite{tsai2018learning}$\dagger$ & 25.6 & 27.2 & 31.8 & 32.1 & 28.8 & 29.2 \\
        MOCDA \cite{gong2021cluster}$\dagger$ & 24.4 & 27.5 & 30.1 & 31.4 & 27.7 & 29.4 \\
        DHA \cite{park2020discover}$\dagger$ & 27.1 & 30.4 & 35.5 & 36.1 & 32.0 & 32.3 \\
        \rowcolor{gray!10} \textit{Ours (AST-OCDA)}$\dagger$ & 32.7 & 32.2 & 38.9 & 39.2 & \textbf{34.6} & \textbf{35.7} \\
    \bottomrule
\end{tabular}}
\vspace{-2mm}
\end{table}

\begin{table}[t]
\centering
\caption{
    Comparison on SYNTHIA$\to$C-Driving benchmark. All methods employ the long-training strategy. $*$ indicates mIoU over 11 classes otherwise over 16 classes.
}
\vspace{-2mm}
\label{tab:synthiatocdriving}
\setlength{\tabcolsep}{1pt}
 \resizebox{1\columnwidth}{!}{
\begin{tabular}{lcccccc}
    \toprule
         \multirow{2}{*}{Method} & \multicolumn{3}{c}{Compound (C)} & \multicolumn{1}{c}{Open (O)} & \multicolumn{2}{c}{Average} \\
        \cmidrule(l{4pt}r{4pt}){2-4} \cmidrule(l{4pt}r{4pt}){5-5} \cmidrule(l{4pt}r{4pt}){6-7}
         & Rainy & Snowy & Cloudy & Overcast & C & C+O \\
        \midrule
        Source only & 16.3 & 18.8 & 19.4 & 19.5 & 18.4 & 18.5 \\
        CBST \cite{zou2018unsupervised} & 16.2 & 19.6 & 20.1 & 20.3 & 18.9 & 19.1 \\
        CRST \cite{zou2019confidence} & 16.3 & 19.9 & 20.3 & 20.5 & 19.1 & 19.3 \\
        ASN \cite{tsai2018learning} & 17.0 & 20.5 & 21.6 & 21.6 & 20.0 & 20.2 \\
        AdvEnt \cite{vu2019advent} & 17.7 & 19.9 & 20.2 & 20.5 & 19.3 & 19.6 \\
        DHA \cite{park2020discover} & 18.8 & 21.2 & 23.6 & 23.6 & 21.5 & 21.8 \\
        \rowcolor{gray!10} \textit{Ours (AST-OCDA)} & 20.3 & 22.6 & 24.9 & 25.4 & \textbf{22.6} & \textbf{23.3} \\
        \midrule
        Source only$*$ & 16.5 & 18.2 & 21.4 & 20.6 & 19.2 & 19.8 \\
        AdvEnt \cite{vu2019advent}$*$ & 21.8 & 22.6 & 26.2 & 25.7 & 23.9 & 24.7 \\
        ASN \cite{tsai2018learning}$*$ & 24.9 & 26.9 & 30.7 & 30.3 & 28.0 & 29.0 \\
        MOCDA \cite{gong2021cluster}$*$ & 26.6 & 27.9 & 32.4 & 31.8 & 29.1 & 30.3 \\
        \rowcolor{gray!10} \textit{Ours (AST-OCDA)}$*$ & 27.9 & 28.8 & 33.9 & 34.2 & \textbf{30.2} & \textbf{31.2} \\
    \bottomrule
\end{tabular}}
\vspace{-2mm}
\end{table}

\begin{table}[t]
\centering
\vspace{-2mm}
\caption{
    Evaluation on extended open domains. mIoU computed over 19 classes for GTA5 as source and 11 classes for SYNTHIA as source, following \citet{gong2021cluster}.
}
\vspace{-2mm}
\label{tab:extendedopeneval}
\setlength{\tabcolsep}{2pt}
 \resizebox{1\columnwidth}{!}{
\begin{tabular}{clcccc}
    \toprule
        \multirow{2}{*}{Source} & \multirow{2}{*}{Method} & \multicolumn{3}{c}{Extended Open} & \multirow{2}{*}{Avg.} \\
        \cmidrule(l{4pt}r{4pt}){3-5}
         &  & Citysc. & KITTI & WildDash &  \\
        \midrule
        \multirow{4}{*}{GTA5} & Source only & 19.3 & 24.1 & 16.0 & 20.5 \\
         & ASN \cite{tsai2018learning} & 22.0 & 23.4 & 17.5 & 22.5 \\
         & MOCDA \cite{gong2021cluster} & 31.1 & 30.9 & 21.6 & 27.8 \\
         & \cellcolor{gray!10}\textit{Ours (AST-OCDA)} & \cellcolor{gray!10}32.6 & \cellcolor{gray!10}31.8 & \cellcolor{gray!10}23.1 & \cellcolor{gray!10}\textbf{29.2} \\
        \midrule
        \multirow{4}{*}{\begin{tabular}{c} SYN- \\ THIA \end{tabular}} & Source only & 24.7 & 20.7 & 17.3 & 20.8 \\
         & ASN \cite{tsai2018learning} & 35.9 & 24.7 & 20.7 & 27.9 \\
         & MOCDA \cite{gong2021cluster} & 32.2 & 34.2 & 25.8 & 31.2 \\
         & \cellcolor{gray!10}\textit{Ours (AST-OCDA)} & \cellcolor{gray!10}37.2 & \cellcolor{gray!10}35.7 & \cellcolor{gray!10}26.9 & \cellcolor{gray!10}\textbf{33.3} \\
    \bottomrule
\end{tabular}}
\end{table}

\begin{table}[t]
\centering
\caption{
    Ablation study on GTA5$\to$C-Driving. \textit{Random} and \textit{Mined} indicate random pairing and hard-mining strategy (Sec \ref{subsubsec:crossmodalsim}\ak{.1}). \textit{Sim} and \textit{Norm} indicate the module usage.
}
\vspace{-2mm}
\label{tab:ablation}
\setlength{\tabcolsep}{2pt}
\resizebox{1\columnwidth}{!}{
\begin{tabular}{cacacacac}
    \toprule
        \multirow{2}{*}{\#} & \cellcolor{white}\multirow{2}{*}{
        \textit{AST-}
        } & \multicolumn{2}{c}{\textit{Pairing}} & \multirow{2}{*}{\textit{AST-}} &  \multicolumn{2}{c}{\textit{Losses}} & \multicolumn{2}{c}{GTA5$\to$C-Driv.} \\
        \cmidrule(l{4pt}r{4pt}){3-4} \cmidrule(l{4pt}r{4pt}){6-7} \cmidrule(l{4pt}r{4pt}){8-9}
         & \cellcolor{white}\textit{Sim} & \textit{Random} & \textit{Mined} & \textit{Norm} & $\mathcal{L}_{s_g}, \mathcal{L}_{t_g}$ & $ \mathcal{L}_{s_g^{\shortarrow}}, \mathcal{L}_{t_g^{\shortarrow}}$ & C & C+O \\
        \midrule
        1. & $\times$ & - & - & $\times$ & - & - & 25.7 & 26.4 \\
        2. & $\times$ & - & - & \checkmark & \checkmark & - & 28.1 & 28.6 \\
        \midrule
        3. & \checkmark & \checkmark & $\times$ & $\times$ & - & - & 27.9 & 28.3 \\
        4. & \checkmark & \checkmark & $\times$ & \checkmark & $\times$ & \checkmark & 30.0 & 30.3 \\
        5. & \checkmark & \checkmark & $\times$ & \checkmark & \checkmark & \checkmark & 31.2 & 31.6 \\
        \midrule
        6. & \checkmark & $\times$ & \checkmark & $\times$ & - & - & 29.0 & 29.8 \\
        7. & \checkmark & $\times$ & \checkmark & \checkmark & $\times$ & \checkmark & 32.8 & 33.6 \\
        8. & \checkmark & $\times$ & \checkmark & \checkmark & \checkmark & \checkmark & \textbf{34.6} & \textbf{35.7} \\
    \bottomrule
\end{tabular}}
\vspace{-2mm}
\end{table}

\begin{figure}[t]
\centering
\vspace{-2mm}
\includegraphics[width=\columnwidth]{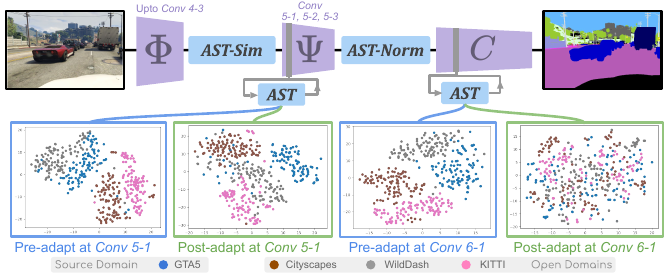} 
\caption{t-SNE plots at the layers following \textit{AST-Sim} and \textit{AST-Norm}, pre- and post-adaptation. Domains remain clustered after \textit{AST-Sim} while they are aligned after \textit{AST-Norm}.}
\label{fig:tsneplots}
\vspace{-4mm}
\end{figure}

\section{Experiments}

We thoroughly evaluate the proposed approach against \textit{state-of-the-art} prior works in the Open Compound DA setting.

\vspace{0.5mm}
\textbf{Datasets.} 
Following \citet{gong2021cluster}, we used the synthetic GTA5 \cite{richter2016playing} and SYNTHIA \cite{ros2016synthia} datasets as the source. In C-Driving \cite{liu2020open}, \textit{rainy}, \textit{snowy}, and \textit{cloudy} sub-targets form the compound target domain, and \textit{overcast} sub-target forms the open domain. Further, we use Cityscapes \cite{cordts2016cityscapes}, KITTI \cite{geiger2013vision} and WildDash \cite{zendel2018wilddash} datasets as extended open domains to test generalization on diverse unseen domains.
All datasets (except SYNTHIA) share 19 semantic categories. We use the mean intersection-over-union (mIoU) metric for evaluating the performance. See Suppl. for more details.

\vspace{0.5mm}
\textbf{Implementation Details.} Following \cite{park2020discover, gong2021cluster}, we employ DeepLabv2 \cite{chen2017deeplab} with a {VGG}16 (\jnkc{Simonyan et al.} \citeyear{simonyan2014very}) backbone as the CNN segmentor. We use SGD optimizer with a learning rate of 1e-4, momentum of 0.9 and a weight decay of 5e-4 during training. We also use a polynomial decay with power 0.9 as the learning rate scheduler. Following \citet{park2020discover}, we use two training schemes for GTA5 \ie, short training scheme with 5k iterations and long training scheme with 150k iterations.
For SYNTHIA, we use only the long training scheme following \citet{gong2021cluster}.

\subsection{Discussion}

We compare our approach with prior arts on GTA5$\to$C-Driving (Table \ref{tab:gta5tocdriving}) and SYNTHIA$\to$C-Driving (Table \ref{tab:synthiatocdriving}) benchmarks as well as on extended open domains (Table \ref{tab:extendedopeneval}). We also provide an extensive ablation study (Table \ref{tab:ablation}).


\vspace{0.5mm}
\textbf{GTA5 as source} (Table \ref{tab:gta5tocdriving}, \ref{tab:extendedopeneval}).
For the short training scheme, our approach outperforms the SOTA method DHA \cite{park2020discover} by an average of 0.7\% on compound domains and 1.2\% on open domain. 
The improvement is enhanced with the long training scheme where we outperform DHA by 2.6\% and 3.1\% on compound and open domains respectively. 
As illustrated in Table \ref{tab:extendedopeneval}, our method outperforms the SOTA method MOCDA \cite{gong2021cluster} on the extended open domains by 1.4\% on average. 
This verifies the better generalization abilities of our approach for new open domains with higher domain-shift from the source data.

\vspace{0.5mm}
\textbf{SYNTHIA as source} (Table \ref{tab:synthiatocdriving}, \ref{tab:extendedopeneval}).
Our approach outperforms the SOTA method DHA \cite{park2020discover} by an average of 1.1\% on compound domains and 1.8\% on the open domain. 
Table \ref{tab:extendedopeneval} shows our superior performance on the extended open domains. We outperform the SOTA \cite{gong2021cluster} with an average 2.1\% over the 3 datasets. This reinforces our superior generalizability.

\vspace{1mm}
\noindent
\textbf{4.1.1 Ablation Study.} 
Table \ref{tab:ablation} presents a detailed ablation to analyze the effectiveness of the proposed components.

First, we evaluate the variations proposed under \textit{AST-Sim}.
As a baseline, we use the source-only performance (\#1).
Deactivating \textit{AST-Norm}
, \textit{AST-Sim} with \textit{random} pairing strategy (\#3 vs. \#1) yields an average 2.1\% improvement.
Next, without \textit{AST-Norm}, \textit{AST-Sim} alongside the proposed \textit{Mined} pairing outperforms the previous (\#6 vs. \#3) by 1.3\%. 
When \textit{AST-Norm} is applied 
alongside only the simulation based losses (\ie $\mathcal{L}_{s_g^{\shortarrow}}, \mathcal{L}_{t_g^{\shortarrow}}$), using \textit{Mined} pairing improves over \textit{random} pairing (\#7 vs. \#4) by 3.0\%. 
Finally, when \textit{AST-Norm} is applied alongside all the four losses, \textit{Mined} pairing outperforms \textit{random} pairing (\#8 vs. \#5) by 3.5\%. This verifies the consistent superiority of our well-thought-of cross-domain pairing strategy (Sec \ref{subsubsec:ast_dn}\ak{.1}).

Second, we evaluate the variations under \textit{AST-Norm}.
Deactivating \textit{AST-Sim}, \textit{AST-Norm} outperforms the baseline (\#2 vs. \#1) by 2.3\% on average. \textit{AST-Norm} alongside simulation-based losses \ie $\mathcal{L}_{s_g^{\shortarrow}}, \mathcal{L}_{t_g^{\shortarrow}}$ (\#4 vs. \#3 and \#7 vs. \#6 for random and mined pairing respectively) yield average improvements of 2.0\% and 3.8\% respectively over not using \textit{AST-Norm}. Next, \textit{AST-Norm} used with all losses improves over the previous (\#5 vs. \#4 and \#8 vs. \#7 for random pairing and hard mining respectively) by 1.2\% and 1.9\% respectively. This verifies the consistent improvement of using all losses over only simulation-based losses.

Both sets of ablations underline the equal importance of \textit{AST-Sim} and \textit{AST-Norm} in the proposed approach.


\vspace{0.5mm}
\noindent
\textbf{4.1.2 Analyzing domain alignment.}
Since \textit{AST-Sim} (layer $l$) and \textit{AST-Norm} (layer $l'$) involve replacing the latent, their effect can be studied at the layer following it.
In Fig. \ref{fig:tsneplots}, we plot t-SNE ({Maaten et al.} \citeyear{tsne}) embeddings of AST-latent at layers $l+1$ and $l'+1$ to examine domain alignment before and after the proposed adaptation. 
Though simulated cross-domain features $(h_{l, s^{\shortarrow}}, h_{l, t^{\shortarrow}})$ span the entire space, projection of the original source and sub-target domains  (\ie $h_{l, s}, h_{l, t}$) stills retains domain-specificity (first and second plots). 
As \textit{AST-Norm} aims to normalize the AST-latent with a fixed mean prototype (\ie enforcing all features to have a common AST-latent), we observe that the domains are aligned post-adaptation (third and fourth plots) at layer $l'+1$. This shows that the proposed \textit{Simulate-then-Normalize} strategy effectively reduces the domain shift in the latent space.

\vspace{-3mm}
\section{Conclusion}

In this work, we performed a thorough analysis of domain discriminability (DDM) to quantify the unwanted correlation between the deep features and domain labels. Based on our observations, we proposed the feature-space Amplitude Spectrum Transformation (AST) to better disentangle and manipulate the domain and task-related information. We provided insights into the usage of AST in two configurations - \textit{AST-Sim} and \textit{AST-Norm} and proposed a novel \textit{Simulate-then-Normalize} strategy for effectively performing Open Compound DA. Extending the proposed approach for more general, realistic DA scenarios such as multi-source, multi-target DA can be a direction for future work.

\noindent
\textbf{Acknowledgements.}
This work was supported by MeitY (Ministry of Electronics and Information Technology) project (No. 4(16)2019-ITEA), Govt. of India.


\appendix


\begin{center}
    \Large
    \textbf{Supplementary Material}
\end{center}

In this supplementary, we provide extensive implementation details with additional qualitative and quantitative performance analysis. Towards reproducible research, we 
will release our complete codebase and trained network weights.

\begin{figure*}[!htbp]
\centering
\includegraphics[width=0.88\textwidth]{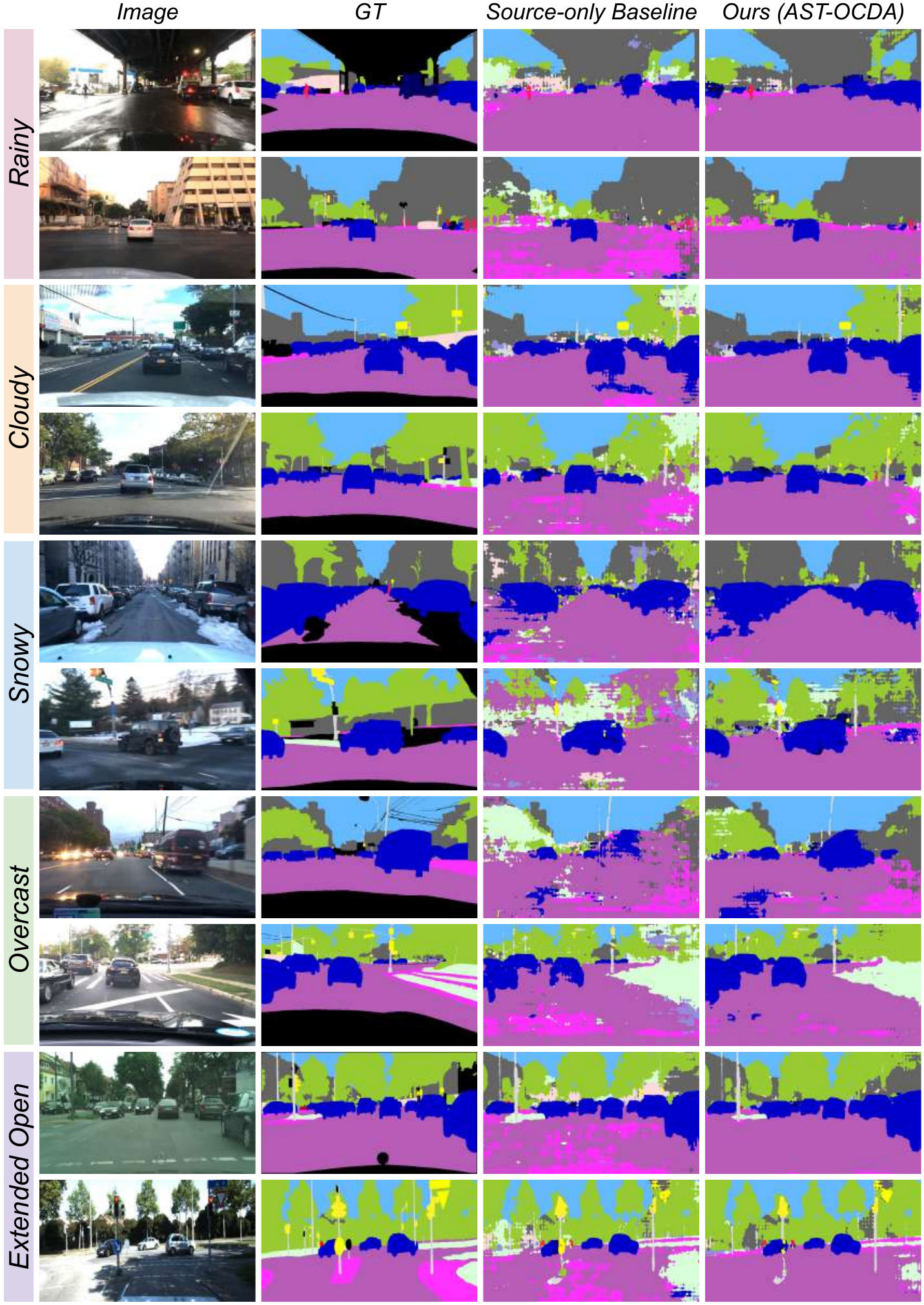} 
\caption{Qualitative analysis of the proposed adaptation for GTA5$\to$C-Driving.}
\label{sup:fig:qualitative}
\end{figure*}

\noindent This supplementary is organized as follows:

\vspace{-1mm}

\begin{itemize}
\setlength{\itemindent}{0mm}
\setlength{\itemsep}{-1mm}
    \item Section~\ref{sup:sec:notations}: Notations (Table~\ref{sup:tab:notations})
    \vspace{1mm}
    \item Section~\ref{sup:sec:implementation}: Implementation Details
    \vspace{-1mm}
    \begin{itemize}
        \setlength{\itemindent}{0mm}
        \item DDM analysis experiments (Sec.~\ref{sup:sec:ddm_analysis}, Fig.~\ref{sup:fig:aug_examples})
        \item Experimental settings (Sec.~\ref{sup:sec:expt_setting}, Table~\ref{sup:tab:ast_enc_dec_arch}, Fig.~\ref{sup:fig:ast_internal})
    \end{itemize}
    \item Section~\ref{sup:sec:analysis}: Analysis
    \vspace{-1mm}
    \begin{itemize}
        \setlength{\itemindent}{0mm}
        \item Quantitative analysis (Sec.~\ref{sup:sec:quantitative}, Table~\ref{tab:synsf_perclass_opentarget}, \ref{tab:gtaperclass_target})
        \item Qualitative analysis (Sec.~\ref{sup:sec:qualitative}, Fig.~\ref{sup:fig:qualitative})
    \end{itemize}
    \vspace{-1mm}
\end{itemize}{}

\section{Notations}
\label{sup:sec:notations}
\noindent We summarize the notations used in the paper in Table \ref{sup:tab:notations}. To enhance the clarity of the approach, we also provide the dimensions of each variable considering $720 \times 1280$ as the input spatial size. The notations are listed under 5 groups viz. Data, Networks, \textit{AST-Sim}, \textit{AST-Norm} and Neural outputs.

\section{Implementation Details}
\label{sup:sec:implementation}

In this section, we provide the extended implementation details excluded from the main paper due to space constraints.

\subsection{DDM analysis experiments}
\label{sup:sec:ddm_analysis}

\begin{figure*}[!htbp]
\centering
\vspace{-3mm}
\includegraphics[width=0.7\textwidth]{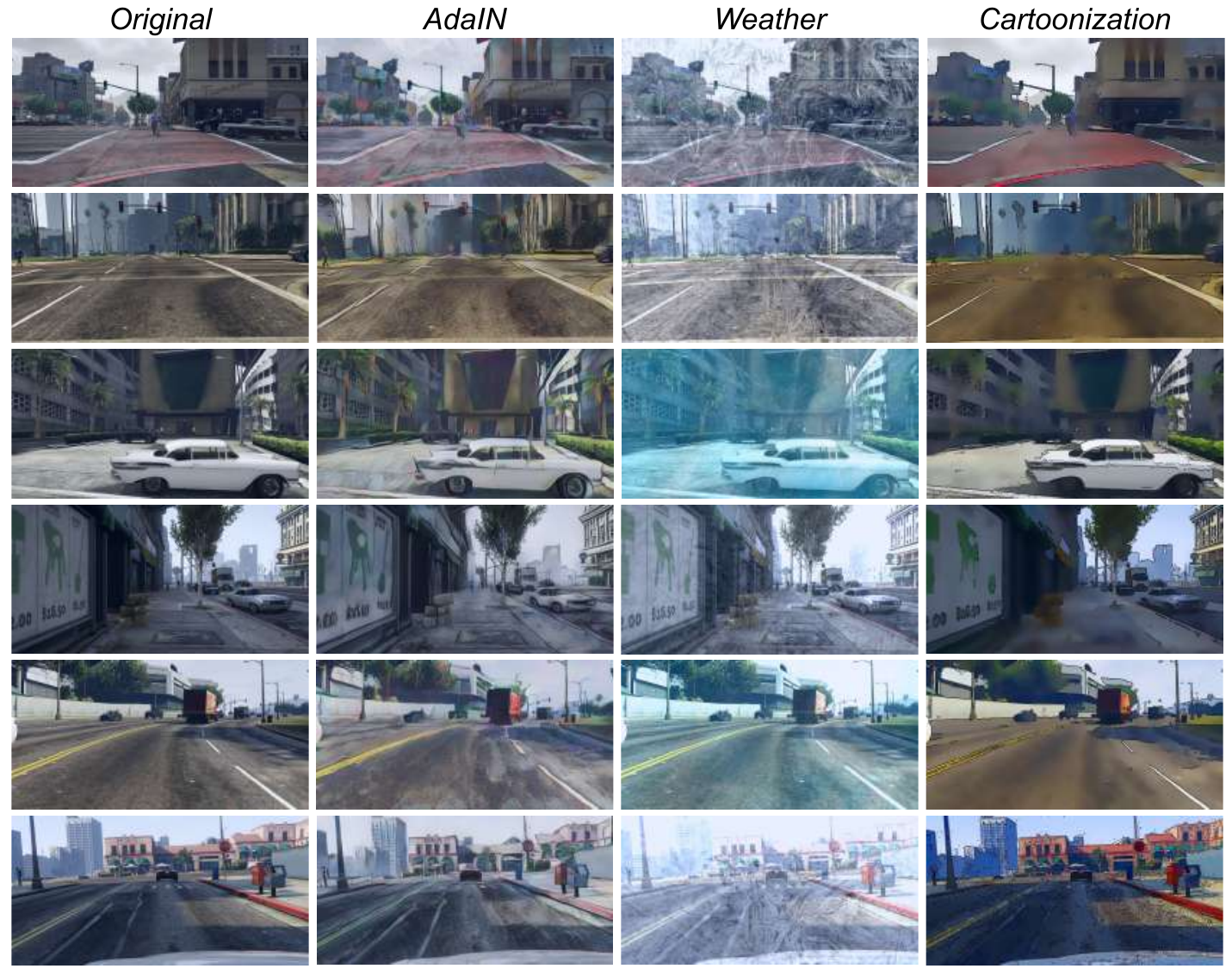} 
\vspace{-3mm}
\caption{Examples of augmentations used for DDM analysis (Sec. \ref{sup:sec:ddm_analysis}). The strong augmentations represent different domains.}
\label{sup:fig:aug_examples}
\end{figure*}

\vspace{2mm}
\textbf{2.1.1 Obtaining multi-domain source.}
We did not use target data for Domain-Discriminability-Metric (DDM) analysis as we intended to use it for OCDA experiments.
We chose to use the following three strong augmentations on source data to represent novel domains (total 4 sub-domains). We show examples of the augmented images in Fig. \ref{sup:fig:aug_examples}.

\vspace{2mm}
\textbf{a) AdaIN}: This technique \cite{huang2017adain} stylizes the image based on a reference image by manipulating the convolutional feature statistics in an instance normalization (\jnkc{Ulyanov et al.} \citeyear{ulyanov2017improved}) layer. We set the strength of stylization to $0.3$ (on a scale of 0 to 1 \ie original image to strongly stylized) to ensure sufficient content preservation.

\vspace{2mm}
\textbf{b) Weather augmentation}: We use the frost (weather condition) augmentation from \citet{imgaug}. There are five levels of severity of which we randomly choose from the lower three levels to control the strength of augmentation.

\vspace{2mm}
\textbf{c) Cartoonization}: We use the cartoonization augmentation from \citet{imgaug}. No controllable parameter is provided for the strength of augmentation.

\vspace{2mm}
\noindent
\textbf{2.1.2 Discriminator training for DDM.}
We train a convolutional discriminator with cross-entropy loss for classifying the input features according to their domain labels. 

\vspace{2mm}
\textbf{a) Disc. for spatial features.} The architecture is similar to that used in \jnkc{Li et al.} (\citeyear{li2019bidirectional}) \ie $5$ conv.\ layers with $4 \times 4$ kernel, stride $2$ and channel numbers $\{64, 128, 256, 512, 4\}$. Here, $4$ channels in the last layer corresponds to the number of sub-domains. Each convolutional layer (except the last) is followed by a LeakyReLU parameterized by $0.2$.

\vspace{2mm}
\textbf{b) Disc. for AST-latent.} We use a fully-connected 2-layer discriminator with 4096 hidden units and ReLU in the first layer and 4 units (no. of sub-domains) in the second layer. 

Despite the lower capacity of the discriminator, DDM of AST-latent is higher than that for convolutional features.
We reported the accuracy of the trained disc. in Fig. \ak{1A} of the main paper for different layer features and AST-latents.

\begin{table}[t]
    \centering
    \caption{\textbf{Notation Table}}
    \setlength{\tabcolsep}{2pt}
    \resizebox{1\columnwidth}{!}{%
        \begin{tabular}{lclc}
        \toprule
            & \multirow{2}{*}{Symbol} & \multirow{2}{*}{Description} & \begin{tabular}{c}
                 Dimensions  \\
                 (C $\times$ H $\times$ W) \\ 
            \end{tabular} \\ 
        \midrule
        \multirow{6}{*}{\rotatebox[origin=c]{90}{Data}} & $\mathcal{D}_s$ & Labeled source dataset & - \\
         & $\mathcal{D}_t$ & Unlabeled target dataset & - \\
         & $\mathcal{D}_t^\textit{(pgt)}$ & Pseudo-labeled target dataset & - \\
         & $(x_s, y_\textit{gt})$ & Labeled source sample & (3, 19) $\times$ 720 $\times$ 1280 \\
         & $x_t$ & Unlabeled target sample & 3 $\times$ 720 $\times$ 1280 \\
         & $(x_t, y_\textit{pgt})$ & Pseudo-labeled target sample & (3, 19) $\times$ 720 $\times$ 1280 \\
        \midrule
        \multirow{5}{*}{\rotatebox[origin=c]{90}{Networks}} & $\Phi$ & CNN (early layers) & - \\
         & $\Psi$ & CNN (middle layers) & - \\
         & $C$ & CNN (dense classifier) & - \\
         & $\texttt{AST}_\textit{cs}$ & \textit{AST-Sim} (cross-domain simulation) & - \\
         & $\texttt{AST}_\textit{dn}$ & \textit{AST-Norm} (domain normalization) & - \\
        \midrule
        \multirow{6}{*}{\rotatebox[origin=c]{90}{\textit{AST-Sim}}} & $h_{l, s}$ & Input source features & $512 \times 90 \times 160$ \\
         & $h_{l, s^{\shortarrow}}$ & Simulated source features & $512 \times 90 \times 160$ \\
         & $z_{l, s}$ & Source \textit{AST-Sim}-latent & $(64 * 512) \times 1 \times 1$ \\
         & $h_{l, t}$ & Input target features & $512 \times 90 \times 160$ \\
         & $h_{l, t^{\shortarrow}}$ & Simulated target features & $512 \times 90 \times 160$ \\
         & $z_{l, t}$ & Target \textit{AST-Sim}-latent & $(64 * 512) \times 1 \times 1$ \\
        \midrule
        \multirow{5}{*}{\rotatebox[origin=c]{90}{\textit{AST-Norm}}} & $h_{l'\!, s}$ & Input orig. source features & $1024 \times 90 \times 160$ \\
         & $h_{l'\!, s^{\shortarrow}}$ & Input simulated source features & $1024 \times 90 \times 160$ \\
         & $h_{l'\!, s_g}$ & Normalized orig. source features & $1024 \times 90 \times 160$ \\
         & $h_{l'\!, s^{\shortarrow}_g}$ & Normalized sim. source features & $1024 \times 90 \times 160$ \\
         & $z_{l'\!, g}$ & Fixed domain prototype & $(64 * 1024) \times 1 \times 1$ \\
        \midrule
        \multirow{6}{*}{\rotatebox[origin=c]{90}{Neural outputs}} & $h_k$ & Layer-$k$ features & $C_k \times H_k \times W_k$ \\ 
         & $z_k$ & Layer-$k$ AST-latent & $(64 * C_k) \times 1 \times 1$ \\
         & $y_{s_g}$ & \textit{AST-Norm} source pred. & $19 \times 720 \times 1280$ \\
         & $y_{s_g^{\shortarrow}}$ & \textit{AST-Sim-then-Norm} source pred. & $19 \times 720 \times 1280$ \\
         & $y_{t_g}$ & \textit{AST-Norm} target pred. & $19 \times 720 \times 1280$ \\
         & $y_{t_g^{\shortarrow}}$ & \textit{AST-Sim-then-Norm} target pred. & $19 \times 720 \times 1280$ \\
        \bottomrule
        \end{tabular}
        }
    \label{sup:tab:notations}
\end{table}

\begin{figure}[t]
\centering
\includegraphics[width=\columnwidth]{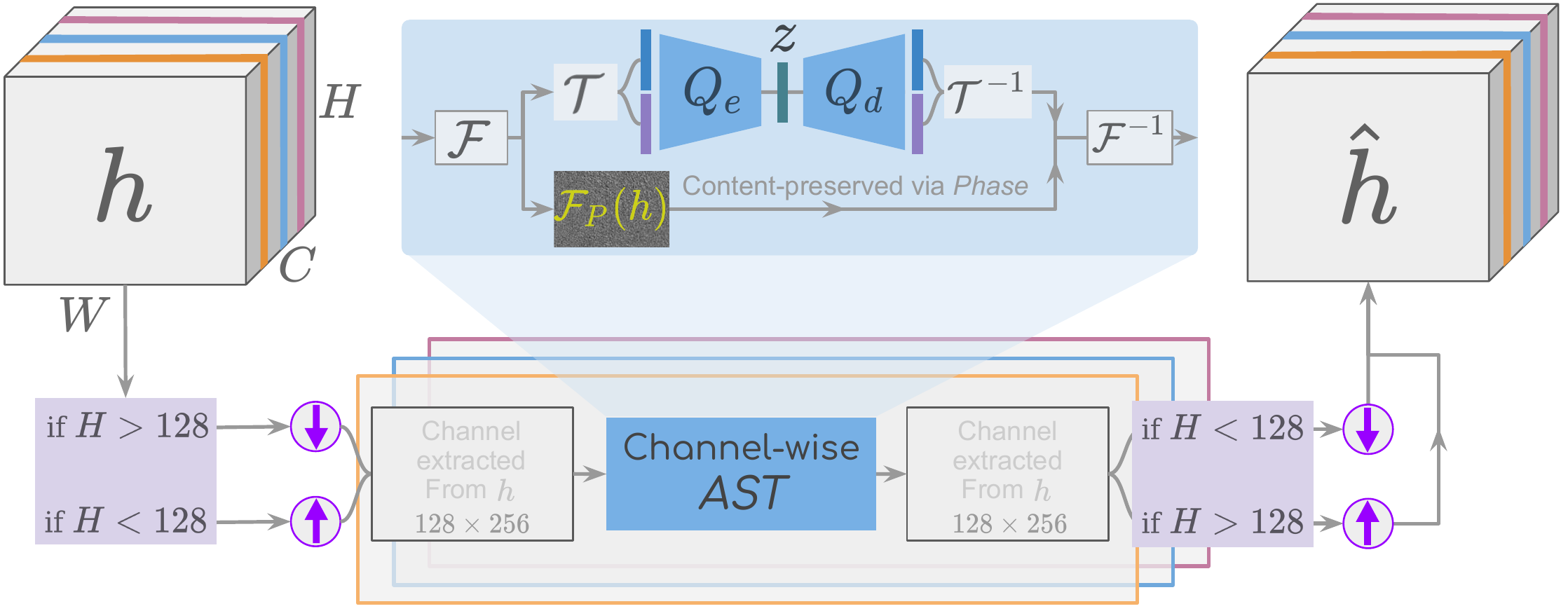} 
\caption{Internal operations involved in AST. Here, purple arrow inside circle indicates spatial upsampling (upward arrow) or downsampling (downward arrow). 
}
\label{sup:fig:ast_internal}
\end{figure}

\begin{table*}[!t]
    \centering
    \caption{\textbf{Quantitative evaluation on SYNTHIA$\rightarrow$C-Driving}.
    Per-class IoU is reported for 11 classes. 
    Compound refers to compound domains \ie rainy, snowy, and cloudy, while Open refers to the open domain from C-Driving \ie overcast. 
    }
    \setlength{\tabcolsep}{9pt}
    \resizebox{\textwidth}{!}{
    \begin{tabular}{llcccccccccccc}
    \toprule
        \multicolumn{14}{c}{SYNTHIA$\rightarrow$C-Driving}\\
        \midrule
         & Method&\rotatebox{90}{road}&\rotatebox{90}{sidewalk}&\rotatebox{90}{building}&\rotatebox{90}{wall}&\rotatebox{90}{fence}&\rotatebox{90}{pole}&\rotatebox{90}{traffic light}&\rotatebox{90}{vegetation}&\rotatebox{90}{sky}&\rotatebox{90}{person}&\rotatebox{90}{car}&mIoU\\
        \midrule
        \multirow{4}{*}{Compound} & 
        Source only & 3.1 & 6.8 & 42.7 & 0.0 & 0.0 & 10.2 & 1.1 & \textbf{39.6} & 69.2 & 9.7 & 28.2 & 19.2\\
        & AdvEnt \cite{vu2019advent} & \textbf{67.2} & 1.8 & 50.7 & 0.0 & 0.0 & 4.4 & 1.3 & 11.7 & 71.8 & 8.7 & 45.7 & 23.9 \\
        & ASN \cite{tsai2018learning} & 63.1 & 11.9 & 46.5 & 0.1 & 0.0 & 10.5 & 3.1 & 22.2 & {78.7} & 17.8 & 54.1 & 28.0\\
        \rowcolor{gray!10} \cellcolor{white!10}&  \textit{Ours (AST-OCDA)} & 65.8 & \textbf{12.4} & \textbf{48.9} & \textbf{0.9} & \textbf{0.4} & \textbf{12.3} & \textbf{5.8} & 26.9 & \textbf{83.5} & \textbf{18.9} & \textbf{56.4} & \textbf{30.2}\\
        \midrule
        \multirow{4}{*}{Open} & 
        Source only & 1.9 & 9.0 & 43.4 & 0.0 & 0.0 & 11.1 & 1.2 & \textbf{45.1} & 74.7 & 13.0 & 27.2 & 20.6 \\
        & AdvEnt \cite{vu2019advent} & 68.9 & 2.5 & 51.6 & 0.0 & 0.0 & 5.7 & 1.4 & 14.2 & 77.2 & 11.7 & 49.3 & 25.7\\
        & ASN \cite{tsai2018learning} & {69.4} & 14.4 & 48.7 & 0.0 & 0.0 & 11.8 & 2.3 & 23.0 & {82.4} & 21.7 & 59.0 & 30.3\\
        \rowcolor{gray!10} \cellcolor{white!10}&  \textit{Ours (AST-OCDA)} & \textbf{73.5} & \textbf{18.6} & \textbf{53.4} & \textbf{0.7} & \textbf{0.3} & \textbf{16.5} & \textbf{8.6} & 27.9 & \textbf{87.5} & \textbf{26.4} & \textbf{63.4} & \textbf{34.2}\\
    \bottomrule
    \end{tabular}
    }
    \label{tab:synsf_perclass_opentarget}
    \vspace{5mm}
\end{table*}

\subsection{Experimental settings}
\label{sup:sec:expt_setting}

Here, we describe the datasets, networks, compute and software used for the experiments.

\vspace{2mm}
\textbf{Datasets.}
The GTA5 dataset \cite{richter2016playing} contains 24966 synthetic road scene images of size extracted from a game engine which are densely labeled for 19 categories. The SYNTHIA dataset \cite{ros2016synthia} contains 2224 synthetic road scene images extracted from a virtual city. The C-Driving dataset \cite{liu2020open} is derived from BDD100k dataset \cite{fisher2020bdd100k}. The training set contains 14697 images from mixed domains without labels while the validation set, with 627 images, is segregated into sub-domains for evaluation purposes. Cityscapes \cite{cordts2016cityscapes} is a real urban road scene dataset collected from European cities. We use 500 images from its validation set for extended open domain evaluation. KITTI \cite{geiger2013vision} consists of road scene images collected from Karlsruhe, Germany. We utilize its validation set with 200 images for extended open evaluation. WildDash \cite{zendel2018wilddash} is a diverse road scene dataset with varying weather, time of day, camera characteristics and data sources. We use its validation set with 70 images for extended open evaluation.

\vspace{2mm}
\textbf{CNN segmentor architecture details.} 
Following \cite{park2020discover, gong2021cluster}, we use DeepLabv2 \cite{chen2017deeplab} with VGG16 (\jnkc{Simonyan et al.} \citeyear{vgg}) backbone as the CNN segmentor. As proposed in Sec \ak{3.4} of the main paper, we divide the CNN segmentor into $\Phi, \Psi,$ and $C$ to accommodate the two AST modules. As shown in Fig. \ak{4} of the main paper, $\Phi$ consists of layers upto \texttt{Conv 4-3}, $\Psi$ consists of layers \texttt{Conv 5-1} to \texttt{Conv 5-3} and $C$ consists of the remaining layers after \texttt{Conv-5-3} of the segmentor.

\vspace{1mm}
\textbf{AST architecture details.}
We illustrate the upsampling/downsampling strategy, for feature map inputs to AST, in Fig. \ref{sup:fig:ast_internal}.
We interpolate all feature map inputs to a spatial size of $128 \times 256$ so that AST can have a fixed capacity encoder $Q_e$ and decoder $Q_d$ irrespective of the layer it is applied at. Further, we interpolate the output features of AST back to the original spatial size.
We show the encoder $Q_e$ and decoder $Q_d$ architecture in Table \ref{sup:tab:ast_enc_dec_arch}. The input and output size of $Q_e$ and $Q_d$ is $128^2 = 16384$ as only half of the amplitude spectrum is required, due to its symmetry about the midpoint (as illustrated in Fig. \ak{2} of the main paper).

\begin{table}[t]
    \centering
    \caption{Encoder-decoder architecture used in AST.}
    \vspace{-3mm}
    \setlength{\tabcolsep}{8pt}
    \resizebox{1\columnwidth}{!}{%
        \begin{tabular}{cccc}
        \toprule
         & Operation & Features & Non-linearity \\
        \midrule
        \multirow{4}{*}{$Q_e$} & Input & 16384 &  \\
         & Fully-connected & 2048 & ReLU \\
         & Fully-connected & 64 & Linear \\
         & Normalization (unit sphere) & 64 &  \\
        \midrule
        \multirow{2}{*}{$Q_d$} & Fully-connected & 2048 & ReLU \\
         & Fully-connected & 16384 & Linear \\
        \bottomrule
        \end{tabular}
        }
    \label{sup:tab:ast_enc_dec_arch}
\end{table}

\vspace{1mm}
\textbf{Compute specifications.} For all experiments, we use a machine with 8-core Intel Core i7-9700F CPU, 64 GB of system RAM and one NVIDIA RTX 8000 GPU (48 GB).

\vspace{1mm}
\textbf{Software specifications.} We use PyTorch 1.8 \cite{paszke2019pytorch} for implementing our proposed method along with Python 3.7 and CUDA 11.0. The 
OS
is Ubuntu 18.04.

\section{Analysis}
\label{sup:sec:analysis}

\vspace{1mm}
\subsection{Quantitative analysis}
\label{sup:sec:quantitative}

We provide extended quantitative results of our proposed approach on SYNTHIA$\to$C-Driving and GTA5$\to$C-Driving benchmarks in Table \ref{tab:synsf_perclass_opentarget}, \ref{tab:gtaperclass_target}. We obtain \textit{state-of-the-art} performance across all benchmarks (Table \ak{2, 3, 4} in main paper).

\vspace{1mm}
\subsection{Qualitative analysis}
\label{sup:sec:qualitative}

We provide qualitative results of our proposed approach against the Source only baseline for the GTA5$\to$C-Driving setting and also on the extended open domains in Fig. \ref{sup:fig:qualitative}. Note that prior arts \cite{park2020discover, gong2021cluster} did not release code or pre-trained models, making it infeasible to compare qualitative results with their methods.

\begin{table*}[!htbp]
    \centering
    \caption{\textbf{Comparison on GTA5$\to$C-Driving}.
    Per-class IoU is reported for 19 classes except `bicycle' class, as it is close to zero for all methods.
    The rainy, snowy and cloudy domains compose the compound target, while overcast is an open domain.
    }
    \vspace{-3mm}
    \setlength{\tabcolsep}{2pt}
    \resizebox{\textwidth}{!}{
    \begin{tabular}{llccccccccccccccccccc}
    \toprule
        \multicolumn{21}{c}{GTA5$\rightarrow$C-Driving (short training strategy following \jnkc{Liu et al.} (\citeyear{liu2020open}))}\\
        \midrule
        &Method&\rotatebox{90}{road}&\rotatebox{90}{sidewalk}&\rotatebox{90}{building}&\rotatebox{90}{wall}&\rotatebox{90}{fence}&\rotatebox{90}{pole}&\rotatebox{90}{traffic light}&\rotatebox{90}{traffic sign}&\rotatebox{90}{vegetation}&\rotatebox{90}{terrian}&\rotatebox{90}{sky}&\rotatebox{90}{person}&\rotatebox{90}{rider}&\rotatebox{90}{car}&\rotatebox{90}{truck}&\rotatebox{90}{bus}&\rotatebox{90}{train}&\rotatebox{90}{motorbike}&mIoU\\
        \midrule
        \multirow{6}{*}{Rainy} & Source only \cite{liu2020open} & 48.3 & 3.4 & 39.7 & 0.6 & 12.2 & 10.1 & {5.6} & 5.1 & 44.3 & 17.4 & 65.4 & 12.1 & 0.4 & 34.5 & 7.2 & 0.1 & 0.0 & 0.5 & 16.2 \\
        & AdaptSegNet \cite{tsai2018learning}& 58.6 & 17.8 & 46.4 & 2.1 & {19.6} & 15.6 & 5.0 & 7.7 & 55.6 & \textbf{20.7} & 65.9 & 17.3 & 0.0 & 41.3 & 7.4 & 3.1 & 0.0 & 0.0 & 20.2 \\
        & CBST \cite{zou2018unsupervised} & 59.4 & 13.2 & 47.2 & 2.4 & 12.1 & 14.1 & 3.5 & 8.6 & 53.8 & 13.1 & \textbf{80.3} & 13.7 & \textbf{17.2} & \textbf{49.9} & 8.9 & 0.0 & 0.0 & {6.6} & 21.3 \\
        & IBN-Net \cite{pan2018IBN} & 58.1 & \textbf{19.5} & 51.0 & 4.3 & 16.9 & \textbf{18.8} & 4.6 & {9.2} & 44.5 & 11.0 & 69.9 & 20.0 & 0.0 & 39.9 & 8.4 & 15.3 & 0.0 & 0.0 & 20.6 \\
        & OCDA \cite{liu2020open} & 63.0 & 15.4 & {54.2} & 2.5 & 16.1 & 16.0 & {5.6} & 5.2 & 54.1 & 14.9 & 75.2 & 18.5 & 0.0 & 43.2 & 9.4 & 24.6 & 0.0 & 0.0 & 22.0 \\
        \rowcolor{gray!10} \cellcolor{white!10}&  \textit{Ours (AST-OCDA)} & \textbf{68.9} & 18.5 & \textbf{59.8} & \textbf{9.7} & \textbf{19.6} & 18.4 & \textbf{12.2} & \textbf{11.3} & \textbf{59.9} & 20.4 & 78.8 & \textbf{25.6} & 3.2 & 48.2 & \textbf{13.8} & \textbf{30.4} & \textbf{0.3} & \textbf{8.5} & \textbf{28.2} \\
        \midrule
        
        \multirow{6}{*}{Snowy} & Source only \cite{liu2020open} & 50.8 & 4.7 & 45.1 & 5.9 & {24.0} & 8.5 & 10.8 & 8.7 & 35.9 & 9.4 & 60.5 & 17.3 & 0.0 & 47.7 & 9.7 & 3.2 & 0.0 & 0.7 & 18.0\\
        & AdaptSegNet \cite{tsai2018learning} & 59.9 & 13.3 & 52.7 & 3.4 & 15.9 & 14.2 & 12.2 & 7.2 & 51.0 & {10.8} & 72.3 & 21.9 & 0.0 & 55.0 & 11.3 & 1.7 & 0.0 & 0.0 & 21.2\\
        & CBST \cite{zou2018unsupervised} & 59.6 & 11.8 & 57.2 & 2.5 & 19.3 & 13.3 & 7.0 & \textbf{9.6} & 41.9 & 7.3 & 70.5 & 18.5 & 0.0 & 61.7 & 8.7 & 1.8 & 0.0 & 0.2 & 20.6\\
        & IBN-Net \cite{pan2018IBN} & 61.3 & \textbf{13.5 }& 57.6 & 3.3 & 14.8 & {17.7} & 10.9 & 6.8 & 39.0 & 6.9 & 71.6 & 22.6 & 0.0 & 56.1 & 13.8 & \textbf{20.4} & 0.0 & 0.0 & 21.9\\
        & OCDA \cite{liu2020open} & 68.0 & 10.9 & 61.0 & 2.3 & 23.4 & 15.8 & 12.3 & 6.9 & 48.1 & 9.9 & 74.3 & 19.5 & 0.0 & 58.7 & 10.0 & 13.8 & 0.0 & 0.1 & 22.9\\
        \rowcolor{gray!10} \cellcolor{white!10}&  \textit{Ours (AST-OCDA)} & \textbf{71.6} & 12.3 & \textbf{65.8} & \textbf{7.5} & \textbf{27.6} & \textbf{18.9} & \textbf{16.4} & 9.4 & \textbf{52.6} & \textbf{13.8} & \textbf{78.9} & \textbf{24.5} & \textbf{1.7} & \textbf{64.8} & \textbf{16.2} & 16.7 & \textbf{0.5} & \textbf{1.7} & \textbf{27.8} \\
        \midrule
        
        \multirow{6}{*}{Cloudy} & Source only \cite{liu2020open} & 47.0 & 8.8 & 33.6 & 4.5 & 20.6 & 11.4 & {13.5} & 8.8 & 55.4 & 25.2 & 78.9 & 20.3 & 0.0 & 53.3 & 10.7 & 4.6 & 0.0 & 0.0 & 20.9\\
        & ASN \cite{tsai2018learning}& 51.8 & 15.7 & 46.0 & 5.4 & 25.8 & 18.0 & 12.0 & 6.4 & 64.4 & 26.4 & 82.9 & 24.9 & 0.0 & 58.4 & 10.5 & 4.4 & 0.0 & 0.0 & 23.8\\
        & CBST \cite{zou2018unsupervised} & 56.8 & 21.5 & 45.9 & 5.7 &  19.5 & 17.2 & 10.3 & 8.6 & 62.2 & 24.3 & {89.4} & 20.0 & 0.0 & 58.0 & 14.6 & 0.1 & 0.0 & 0.1 & 23.9\\
        & IBN-Net \cite{pan2018IBN} & 60.8 & 18.1 & 50.5 & 8.2 & 25.6 & {20.4} & 12.0 & \textbf{11.3} & 59.3 & 24.7 & 84.8 & 24.1 & \textbf{12.1} & 59.3 & 13.7 & 9.0 & 0.0 & 1.2 & 26.1\\
        & OCDA \cite{liu2020open} & 69.3 & 20.1 & 55.3 & 7.3 & 24.2 & 18.3 & 12.0 & 7.9 & 64.2 & {27.4} & 88.2 & 24.7 & 0.0 & 62.8 & 13.6 & 18.2 & 0.0 & 0.0 & 27.0\\
        \rowcolor{gray!10} \cellcolor{white!10}&  \textit{Ours (AST-OCDA)} & \textbf{72.5} & \textbf{24.6} & \textbf{58.1} & \textbf{10.8} & \textbf{27.6} & \textbf{22.3} & \textbf{14.5} & 10.2 & \textbf{67.8} & \textbf{29.5} & \textbf{90.6} & \textbf{28.6} & 2.5 & \textbf{64.6} & \textbf{18.9} & \textbf{23.7} & \textbf{0.4} & \textbf{2.2} & \textbf{31.6} \\
        \midrule
        
        \multirow{6}{*}{Overcast} & Source only \cite{liu2020open} & 46.6 & 9.5 & 38.5 &  2.7 & 19.8  & 12.9 & {9.2} & 17.5 &  52.7 &  19.9 & 76.8 &  20.9 & 1.4 & 53.8 & 10.8 & 8.4 & 0.0 & 1.8 & 21.2\\
        & ASN \cite{tsai2018learning}& 59.5 & 24.0 & 49.4 &  6.3 & 23.3 & 19.8 & 8.0 & 14.4 & 61.5 & 22.9 & 74.8 & 29.9 & 0.3 & 59.8 &  12.8 & 9.7 & 0.0 & 0.0 & 25.1 \\
        & CBST \cite{zou2018unsupervised} & 58.9 & 26.8 & 51.6 & 6.5 & 17.8 & 17.9 & 5.9 & {17.9} & 60.9 & 21.7 & {87.9} & 22.9 & 0.0 & 59.9 & 11.0 & 2.1 & 0.0 & 0.2 & 24.7\\
        & IBN-Net \cite{pan2018IBN} & 62.9 & 25.3 & 55.5 & 6.5 &  21.2 & {22.3} & 7.2 & 15.3 & 53.3 & 16.5 & 81.6 & 31.1 & 2.4 & 59.1 & 10.3 & 14.2 & 0.0 & 0.0 & 25.5\\
        & OCDA \cite{liu2020open} & 73.5 & 26.5 & 62.5 & 8.6 & 24.2 & 20.2 & 8.5 & 15.2 & 61.2 & 23.0 & 86.3 & 27.3 & 0.0 & 64.4 & 14.3 & 13.3 & 0.0 & 0.0 & 27.9\\
        \rowcolor{gray!10} \cellcolor{white!10}&  \textit{Ours (AST-OCDA)} & \textbf{77.6} & \textbf{29.7} & \textbf{67.8} & \textbf{12.7} & \textbf{26.8} & \textbf{25.6} & \textbf{13.8} & \textbf{19.5} & \textbf{67.3} & \textbf{27.5} & \textbf{90.5} & \textbf{34.2} & \textbf{4.2} & \textbf{72.8} & \textbf{19.6} & \textbf{18.4} & \textbf{0.6} & \textbf{3.5} & \textbf{34.0} \\
   \bottomrule
    \end{tabular}
    }
    \label{tab:gtaperclass_target}
\end{table*}

\pagebreak

{
\small
  \bibliography{aaai22}
}

\end{document}